# UNMASKING HONEY ADULTERATION: A BREAKTHROUGH IN QUALITY ASSURANCE THROUGH CUTTING-EDGE CONVOLUTIONAL NEURAL NETWORK ANALYSIS OF THERMAL IMAGES

Ilias Boulbarj[1], Bouklouze Abdelaziz[2], Yousra El Alami[2], Douzi Samira[3] and Douzi Hassan[1]

[1] IRF-SIC Laboratory, Ibn Zohr University's -Agadir ,Morocco
h.douzi@uiz.ac.ma
boulbarjilias@gmail.com

[2] Biopharmaceutical and Toxicological Analysis Research Team, Laboratory of Pharmacology and Toxicology, Faculty of Medicine and Pharmacy, Mohammed V University in Rabat, Morocco
a.bouklouze@um5r.ac.ma
usralami@gmail.com

[3] Faculty of Medicine and Pharmacy, Mohammed V University in Rabat, Morocco
s.douzi@unm5r.ac.ma

## ABSTRACT

*Honey, a natural product generated from organic sources, is widely recognized for its revered reputation. Nevertheless, honey is susceptible to adulteration, a situation that has substantial consequences for both the well-being of the general population and the financial well-being of a country. Conventional approaches for detecting honey adulteration are often associated with extensive time requirements and restricted sensitivity. This paper presents a novel approach to address the aforementioned issue by employing Convolutional Neural Networks (CNNs) for the classification of honey samples based on thermal images.*

*The use of thermal imaging technique offers a significant advantage in detecting adulterants, as it can reveal differences in temperature in honey samples caused by variations in sugar composition, moisture levels, and other substances used for adulteration. To establish a meticulous approach to categorizing honey, a thorough dataset comprising thermal images of authentic and tainted honey samples was collected. Several state-of-the-art Convolutional Neural Network (CNN) models were trained and optimized using the dataset that was gathered. Within this set of models, there exist pre-trained models such as InceptionV3, Xception, VGG19, and ResNet that have exhibited exceptional performance, achieving classification accuracies ranging from 88% to 98%. Furthermore, we have implemented a more streamlined and less complex convolutional neural network (CNN) model, outperforming comparable models with an outstanding accuracy rate of 99%. This simplification offers not only the sole advantage of the model, but it also concurrently offers a more efficient solution in terms of resources and time. This approach offers a viable way to implement quality control measures in the honey business, so guaranteeing the genuineness and safety of this valuable organic commodity.*

## 1. Introduction

Honey is a palatable and nourishing substance that has been consumed globally for millennia. In addition to imparting a pleasant taste to our culinary preparations, this substance is also abundant in essential elements that confer advantageous effects on human well-being. Honey is a saccharine and thick material that is synthesized by bees through the transformation of plant nectar [1]. The latter refers to a naturally occurring food that is abundant in essential nutrients. This substance comprises essential vitamins, minerals, antioxidants, and enzymes that provide advantageous effects on human health. Honey is abundant in naturally occurring sugars, namely fructose and glucose, which serve as readily available energy sources for the human body. Additionally, this substance possesses healing, antioxidant, antibacterial, and anti-inflammatory characteristics, which have the potential to aid in the treatment of many skin diseases and infections [2]. The utilization of this substance by ancient Egyptians was primarily for the purpose of preserving deceased bodies through embalming practices, whilst the Greeks and Romans employed it as a means to expedite the healing process of wounds [3]. In contemporary medical practice, honey has emerged as a commonly employed therapeutic agent for the management of burns, wounds, and ulcers.

The fall in the worldwide bee population can be attributed to the adverse effects of environmental degradation and the transfer of diseases [4]. The previously described issue has led to an increase in the demand for honey, thereby resulting in a scarcity of honey supply. As a result, it becomes apparent that this particular food item is highly vulnerable to fraudulent practices [5]. Unfortunately, there has been a rise in instances of food fraud, particularly targeting honey, as a means to meet the growing demand [6]. Fraud can emerge in either a direct or indirect manner [4]. In cases of deliberate adulteration, some ingredients, such as artificial sweeteners, glucose syrup, inverted sugar, cane sugar syrup, and beet sugar [1], are purposefully added to honey in order to increase its volume or alter its flavor. The term "indirect adulteration" pertains to a procedure wherein bees are supplied with a substance with the intention of contaminating the honey they generate, hence leading to the inclusion of this component in the ultimate honey product.

The potential health consequences of consuming adulterated honey are a matter of concern. Artificial sweeteners, such as glucose syrup and inverted sugar, are commonly utilized as alternatives to unadulterated honey. The use of these artificial sweeteners has been linked to numerous health issues, encompassing but not restricted to obesity, diabetes, and cardiovascular disease. Moreover, the act of adulterating honey possesses the capacity to reduce the number of beneficial elements that contribute to human well-being. The inclusion of some additional substances in products can have significant health consequences for customers who have allergies or intolerances to those ingredients. Moreover, the presence of food fraud in the honey sector may have negative implications for the economy, as it reduces the value and quality of authentic products, so impacting the activities of legitimate producers.

Computer vision is an extensively employed field of deep learning that encompasses the application of algorithms for the purpose of identifying and categorizing diverse components inside visual input, such as objects, humans, and other entities. A considerable body of academic literature has been dedicated to investigating the utilization of computer vision techniques for the purpose of detecting instances of adulteration or fraudulent activities in the realm of food goods. The utilization of a Convolutional Neural Network (CNN) model, as outlined in the scholarly source [7], has been employed to detect occurrences of fraudulent coffee grain. The detection procedure entails the instantaneous categorization of two discrete coffee cultivars. The researchers employed the computer vision model described in reference [8] to identify the presence of papaya seeds, which are frequently used as an adulterant, within samples of black pepper. In an independent investigation, researchers utilized a sophisticated convolutional neural network (CNN) to classify images of turmeric powder in order to detect fraudulent practices [9]. The researchers introduced a novel methodology in their investigation, which integrates computer vision technology and an electronic nose for the purpose of identifying cases of saffron adulteration [10]. In addition, scholars have undertaken inquiries into the application of computer vision technology in the agricultural sector, with a particular emphasis on the detection and identification of diseases and pests that impact tomato crops [11].

This research presents a methodology that utilizes an original data set comprising thermal pictures of both pure and contaminated honey samples, which have different concentrations. Thermal images are acquired through the utilization of a thermal camera. The dataset is used to train advanced Convolutional Neural Network (CNN) models that can accurately detect tainted samples. In addition to these models, a Convolutional Neural Network (CNN) model has been introduced. This model is distinguished by its simplicity and has demonstrated superior performance compared to other pre-trained models in distinguishing adulterated samples from non-adulterated ones.

The following delineates the key contributions of this research article:

- This work presents a new dataset comprising thermal pictures of both pure and tainted honey samples.

- We conduct the identification of significant sections inside each image. This method involves identifying and accurately outlining important areas within each image, while also ensuring that all significant information is preserved.

- We introduce a novel method called "Temperature Fluctuations" to enhance the quality of our thermal dataset. This method entails introducing stochastic alterations to the pixel values of thermal images. Our objective is to enhance the diversity of our dataset and enhance the resilience of models trained on this dataset.

- A battery of tests was performed to evaluate the efficacy of various state-of-the-art Convolutional Neural Network (CNN) models in identifying genuine and adulterated honey sample images.

- This paper also introduces a Convolutional Neural Network (CNN) model. This model is distinguished by its simplicity and has demonstrated outstanding performance in distinguishing adulterated samples from non-adulterated ones, outperforming other pre-trained models.

## 2. Materials and methods

### 2.1. Honey's Samples Preparation

A set of 24 samples of thyme honey, exhibiting diverse weight percentages of glucose ranging from 10% to 50%, were meticulously produced for the purpose of examination. The experimental procedure involved the utilization of an analytical balance. The complete dataset consisted of a total of 10 unadulterated honey samples, in addition to 5 samples of honey contaminated with 10%, 25%, and 4 samples with 50% adulteration. The samples were placed into a temperature-controlled oven set at 60 °C. The specimens were placed into transparent plastic cuvettes specifically constructed for spectroscopic examination, with a path length of 5 cm. The samples were incubated within the cuvettes for a duration of 24 hours prior to measurement.

Table 1. The distribution of samples according to their respective percentages of adulteration.

| Levels of Adulteration | 0% | 10% | 25% | 50% | Total |
|---|---|---|---|---|---|
| Number of samples | 10 | 5 | 5 | 4 | 24 |

### 2.2. Thermal Image Acquisition

The present study employs a thermal camera to capture the thermal radiation emitted by honey while it performs the cooling process under warm conditions. The research employed the FLIR ONE PRO thermographic camera, which has a spectral range that extends from 8 to 14µm. The device has a precision rate of 3%, a frame rate of 8.7 Hz, a thermal resolution of 160 x 120 pixels, and possesses the ability to detect temperatures within the range of 20 to 400 degrees Celsius [12]. The camera being examined is a portable device that is outfitted with a USB-C interface, enabling its connectivity to a smartphone or tablet. A collection of 15-minute films was captured for every honey sample. The thermal camera is utilized to visually represent the spatial distribution of thermal energy on a particular surface. It possesses the ability to fuse thermal images with visible images, hence improving the identification of heat sources (Figure 1).

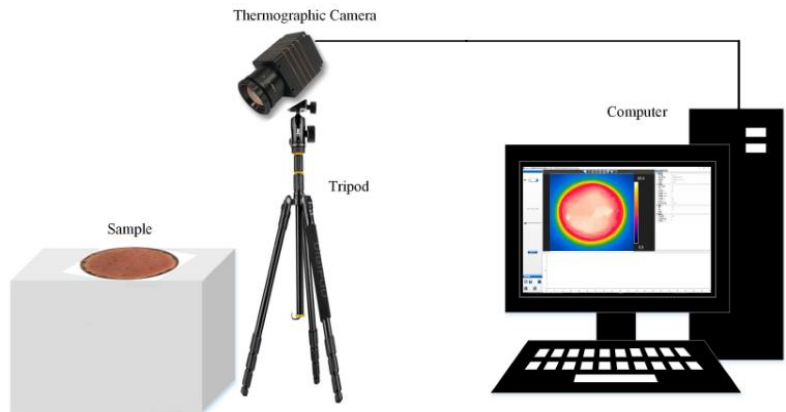

Figure 1. Protocol of experiments

### 2.3. Thermal images preprocessing

#### 2.3.1. Thermal image extraction

The extraction of images from films is achieved by the utilization of the FFMPEG software [13]. The latter is utilized for the alteration, conversion, and processing of a diverse range of media assets, including videos. The FFMPEG software was employed to extract frames from the recorded videos of the honey samples. Specifically, a frame was extracted at a frequency of one frame per thirty seconds during the frame extraction procedure. The outcome of this process will yield a database including 360 frames. The subsequent table (Table 2.) provides a complete analysis of the distribution of the database.

Table 2. Displays the number of thermal images that were extracted based on their respective degrees of adulteration.

| Levels of Adulteration | 0% | 10% | 25% | 50% | Total |
|---|---|---|---|---|---|
| Number of extracted thermal images | 150 | 75 | 75 | 60 | 360 |

The primary objective of this study is to classify honey samples into two distinct categories: Adulterated and Not Adulterated. The study is based on the examination of 150 thermal images acquired from 10 unadulterated honey samples, in addition to the scrutiny of 210 thermal images from 14 tainted samples. The latter samples exhibit different degrees of adulteration, with amounts of 10%, 25%, and 50% correspondingly (Table 3.). The range of adulteration levels serves as a thorough assessment of our approach's capacity to distinguish between samples with varying degrees of adulteration and samples of pure honey.

Table 3. Displays the number of adulterated and unadulterated thermal images.

|  | Unadulterated | Adulterated | Total |
|---|---|---|---|
| **Number of thermal images** | 150 | 210 | 360 |

### 2.3.2. ROI Detection

After the completion of the extraction process, a second phase known as ROI Detection [14] or the identification of areas of significance is conducted. The suggested methodology involves the identification and accurate demarcation of crucial areas within each image, while concurrently ensuring the preservation of all pertinent information. This process enables the reduction of noise and the elimination of less significant information. The employed methodology encompassed the application of image processing algorithms, such as edge detection, picture segmentation, and machine learning-based techniques, to identify and separate relevant regions. The approach described above begins with an initial pre-processing stage in which the original image is converted into a grayscale version. The objective of this alteration is to improve the capacity to distinguish the boundaries of objects inside the image for subsequent analysis. Following the acquisition of the grayscale image, an edge detection technique is subsequently employed. The aforementioned method has been specifically developed to identify and examine significant fluctuations in luminosity within an image, hence accentuating the delineations and contours of the shown objects. The result of the contour detection procedure produces a mask, which is then utilized to superimpose over the original image. The utilization of this mask leads to the creation of a unique image, commonly known as a masked image. The masked image effectively retains relevant data while removing unnecessary components, as illustrated in Figure 2. In order to optimize the utilization of computational resources and minimize execution time, the dimensions of the photographs have been downscaled to a resolution of 300 by 300 pixels. The manipulation of image pixels is facilitated through the utilization of various Python tools, including OpenCV [15].

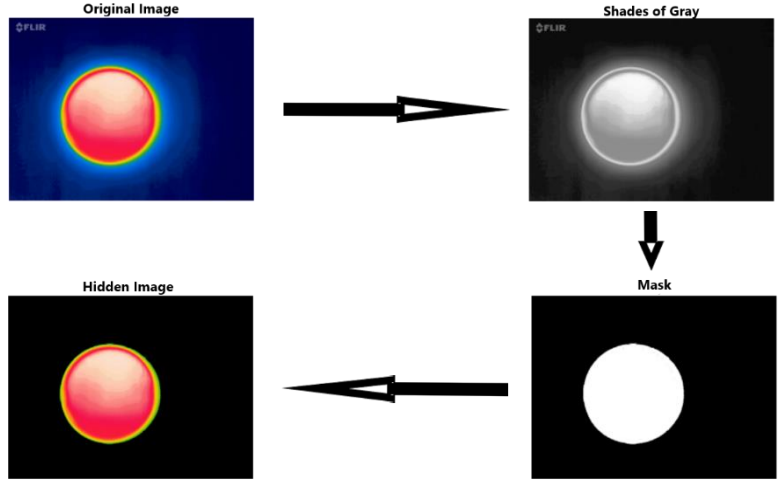

Figure 2. Region of Importance Detection Process

### 2.4. Data Augmentation

In order to enhance the diversity of our thermal dataset and enhance the resilience of models trained on this dataset, we suggest the implementation of a novel augmentation technique referred to as "Temperature Fluctuations". The fundamental concept of the Temperature Fluctuations approach entails the incorporation of random perturbations into the pixel intensities of thermal pictures. This process generates artificial variations in temperature, so simulating a wide range of environmental conditions. The objective of this phase is to enhance the models' exposure to a broader spectrum of temperature models, hence enhancing their capacity to generalize more efficiently. The methodology under consideration is purposefully engineered to process thermal image inputs within a specified temperature range, with the ultimate objective of augmentation. The stochastic generation of temperature fluctuations occurs within the designated range, after which they are superimposed onto the pixel intensities of the given image.

Let I symbolize the initial thermal image, $T_{variation}$ represent the temperature variation created randomly, and $I_{augmented}$ designate the resulting augmented image. The process of augmentation can be properly represented in a systematic manner.

$$I_{augmented} = clip(I + T_{variation}, 0, 255)$$

The clip function is utilized to constrain pixel values inside the permissible temperature range, which spans from 0 to 255. The use of intentionally induced temperature variations can improve the effectiveness and robustness of models trained on enhanced data, allowing them to effectively handle a broader spectrum of thermal circumstances (Figure 3.). The proposed methodology demonstrates attributes of being lightweight, readily implementable, and provides an additional tool to enhance the credibility and diversity of our thermal datasets.

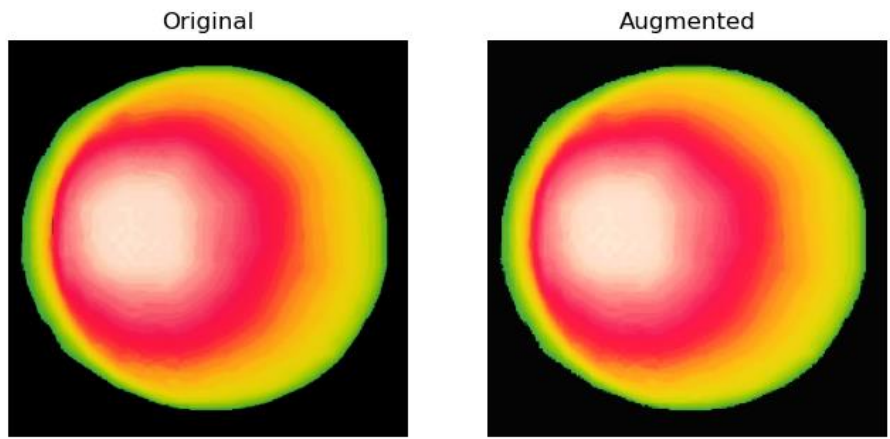

Figure 3. Thermal image prior and after to the implementation of temperature fluctuations

Color images typically consist of three channels: red, green, and blue (RGB). The values in each channel represent the intensity of the corresponding Color. To visually represent the extent of change between the original and augmented images, we compute the absolute disparity for each channel separately to graphically depict the magnitude of change in each Color channel between the original and generated images (Figure 4.). The absolute difference image exhibits a higher magnitude for a particular channel when there is a more noticeable variation in intensity for that Color in the corresponding region of the image. Brighter areas indicate locations where the pixel values have experienced more significant alterations between the original and generated image, whereas colored portions show regions where the pixel values have had less significant alterations.

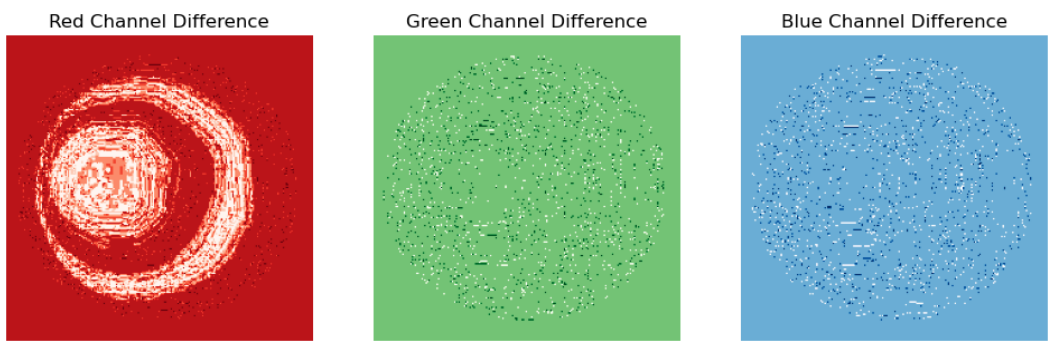

Figure 4. displays the absolute difference for each channel between the original and generated image seen in Figure 3.

In this work, we utilized this augmentation strategy to enhance the magnitude of our dataset. In order to accomplish this, we introduced random perturbations within the range of [-5,5] to modify the pixel intensities of each image. As a result, we obtained an additional 360 images (Table 4) that would significantly expand the diversity of our thermal dataset. Consequently, models trained on this dataset can improve their classification capabilities.

Table 4. Displays the number of adulterated and unadulterated thermal images after augmentation.

|  | **Unadulterated** | **Adulterated** | **Total** |
|---|---|---|---|
| **Number of thermal images** | 300 | 420 | 720 |

### 2.5. Models' Development

#### 2.5.1. CNN architecture

Convolutional Neural Networks (CNNs) are a type of neural network algorithms distinguished by their multi-layered structure. The neural network architecture comprises multiple layers, starting with an initial input layer that receives input images. This is subsequently followed by a sequence of convolution layers and pooling layers [16]. The layers inside the network are responsible for performing non-linear transformations on the input data, hence enhancing the network's capacity to identify and analyze patterns and structures present in the data. In the end, a fully connected layer is utilized to generate a probability vector. This vector is responsible for indicating the relative probabilities of the images being associated with specific classes, as depicted in Figure 5.

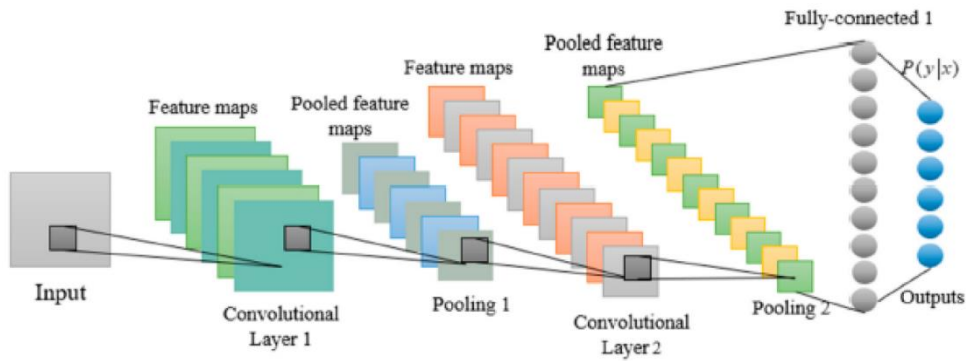

Figure 5. Architecture of a convolutional neural network

#### 2.5.2. Pretrained Convolutional Neural Network

A pretrained Convolutional Neural Network (CNN) denotes a model that has undergone training on an extensive dataset tailored for a particular task, resulting in the preservation of its learnt parameters, namely weights and biases. The utilization of this preexisting model can serve as an initial foundation for a subsequent task that is closely associated, eliminating the need for starting the training process over. The utilization of transfer learning is especially advantageous in cases where the target job possesses a scarcity of labeled data, as the pretrained model has already acquired generic features from an extensive dataset.

Within the realm of widely acknowledged Pretrained CNNs, there exist multiple instances, including VGG [17], ResNet [18], Inception [19], and Xception [20] ... etc. Each model serves to construct an architectural framework that outlines the network's structure, as well as the arrangement of its layers and processes.

The acquisition of weights and biases for each pre-trained model is accomplished through the training process. The utilization of these parameters aims to capture unique attributes and recognizable patterns within the input data, equipping the model with the capability to discern and classify things according to specified categorizations.

This study employed four pre-trained CNNs architectures, specifically VGG [17], ResNet [18], Inception [19], and Xception [20].

### 2.5.2.1. Inception Net model

The Inception Net model, commonly referred to as GoogleNet [19], is an innovative and groundbreaking approach in the realm of deep learning created by Google's research team (Figure 6.). The primary objective of its development was to optimize the efficiency of neural networks by minimizing parameters and enhancing accuracy in classification. The architectural design comprises multiple parallel branches that perform convolution operations with different kernel sizes, allowing the architecture to efficiently capture patterns at various spatial scales. The main parallel branch of the inception module utilizes a 1x1 convolution approach to efficiently reduce the number of activation channels. The convolutional layer is highly valuable because to its capacity to efficiently reduce the number of parameters in the neural network. The second branch includes a convolutional layer with a 3x3 kernel size, which allows for the detection of complex patterns with improved precision and accuracy. The third branch utilizes a convolutional operation with a kernel size of 5x5, which allows for the identification of patterns over a wider spatial area. In order to accurately capture worldwide trends, the fourth branch utilizes a pooling operation with a kernel size of 3x3, as depicted in Figure 6. The output produced by multiple simultaneous branches is combined to create a cohesive output, which is then transferred to the next layer. InceptionNet has emerged as a robust alternative for image classification problems, showcasing exceptional performance with reduced complexity in comparison to other modern designs. The combination of its capacity to uphold superior classification accuracy while efficiently utilizing CPU resources renders it a compelling option for practical applications, particularly within the realm of computer vision and deep learning.

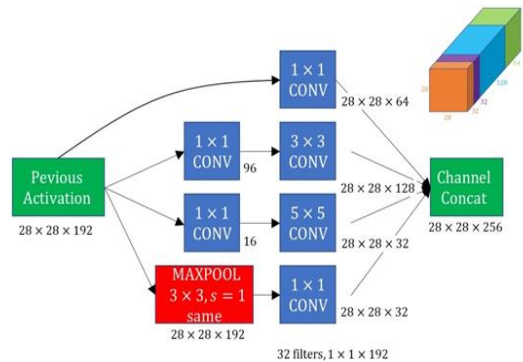

Figure 6. Depicts the architectural designs of Inception module.

### 2.5.2.2. ResNet model

The ResNet (Figure 7.), also known as the Residual Network [18], is a convolutional neural network structure designed to overcome the difficulties faced when training deep neural networks with multiple layers. It was introduced by Kaiming He et al. [17]. ResNet models utilize a distinctive learning approach that focuses on the concept of residual mapping. This idea involves finding the specific changes needed in the input in order to achieve the intended outcome. The attainment of this is facilitated by the utilization of skip connections, which are alternatively referred to as shortcut connections or identity mappings. Skip connections are a mechanism that facilitates the propagation of gradients during training, leading to a smoother flow. When using ResNet designs, it is common to include Global Average Pooling in the last layer instead of using fully connected layers. The purpose of employing the Global Average Pooling (GAP) operation is to condense the spatial dimensions of the feature maps into a singular value for each channel. The ResNet topologies exhibit a range of depths, as seen in the influential study where ResNet models with 34, 50, 101, and 152 layers were first introduced.

The ResNet architecture has significantly transformed the field of neural networks with the introduction of a groundbreaking idea known as residual connections [18]. ResNet has proven to be a successful approach in addressing the difficulties associated with training deep networks and surpassing the constraints of earlier architectures. ResNet is frequently utilized for intricate tasks like image segmentation and object recognition due to its capacity to acquire profound hierarchical representations. Due to its modular construction and capacity to sustain optimal performance even with highly complex networks, it is widely favored in the realm of computer vision.

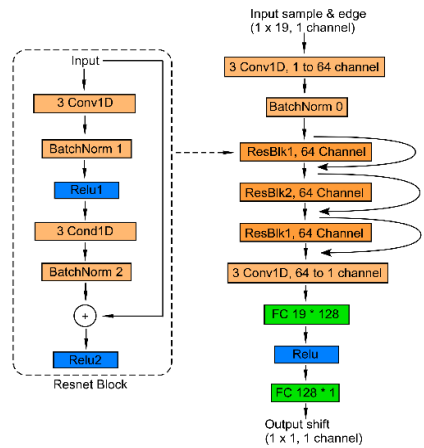

Figure 7. Depicts the architectural designs of ResNet module.

### 2.5.2.3. VGG19

The VGG19 architecture is a convolutional neural network (CNN) developed by the Visual Geometry Group at the University of Oxford (Figure 8.). The research paper titled "Very Deep Convolutional Networks for Large-Scale Image Recognition" introduced the notion [17].

The VGG19 model has a consistent and homogeneous structure, comprising a total of 19 layers, with 16 levels specifically designed for convolutional operations and 3 fully linked layers. Convolutional layers employ 3x3 filters with a stride of 1, in alignment with a sequence of successive convolutional layers succeeded by max pooling layers. The incremental stacking of minuscule filters enables the neural network to acquire intricate characteristics.

The primary objective behind the development of the VGG19 model was to fulfill the demands of complex image recognition jobs. The model demonstrated a highly competitive performance on the ImageNet Large Scale Visual Recognition Challenge (ILSVRC) dataset. Although VGG19 may not be as computationally efficient as more recent architectures, it is nonetheless extensively utilized for transfer learning and serves as a fundamental benchmark in numerous computer vision applications.

The VGG19 architecture is a significant improvement in the field of convolutional neural networks (CNN). It stands out for its consistent design, which simplifies the architecture and enhances its comprehensibility. Additionally, the inclusion of fully connected layers in the final stage of the network enables VGG19 to effectively identify intricate patterns in the input data [17].

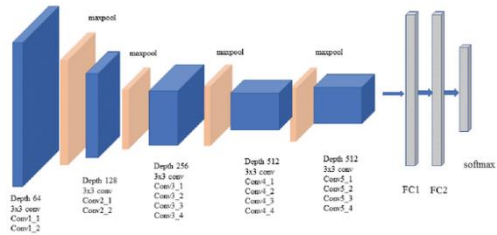

Figure 8. Depicts the architectural designs of VGG19.

### 2.5.2.4. Xception model

The Xception architecture, sometimes known as "Extreme Inception", is a significant advancement of the inception concept originally offered by GoogleNet. Xception adopts a significantly distinct strategy by substituting conventional convolution blocks with deep separable ones (see to Figure 9). This method of dividing convolution procedures into two separate steps [20] - one for spatial convolution and another for deep convolution - effectively decreases the quantity of model parameters while maintaining the capability to extract intricate features. The Xception architecture, like many contemporary models, employs Global Average Pooling (GAP) in its last layer instead of using fully connected layers. This architectural design enables improved generalization across diverse datasets while minimizing the danger of overfitting. Xception demonstrated exceptional performance across a range of tasks, including image classification and semantic segmentation. The Xception approach's innovation renders it a compelling option for achieving an optimal equilibrium between performance power and computing efficiency.

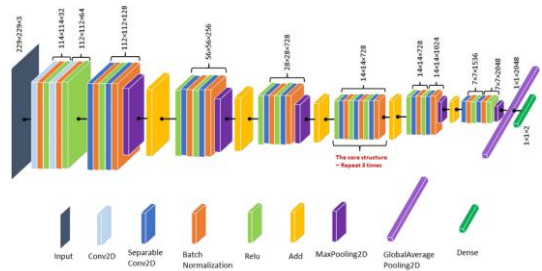

Figure 9. Depicts the architectural designs of Xception.

These state-of-the-art convolutional neural network (CNN) architectures were utilized in the present study; they were trained and optimized using a previously published dataset. The primary aim of this procedure was to ascertain the presence or absence of adulteration in the samples.

### 2.5.2.5. Proposed CNN Model

In addition to using pre-trained models, a new Convolutional Neural Network (CNN) structure is created by combining five Convolutional Layers (Conv2D) with 18, 18, 32, 64, and 128 filters respectively. Each layer uses a (3, 3) filter size and ReLU activation. MaxPooling Layers

(MaxPooling2D) are applied after each convolutional layer, and Batch Normalization Layers [22] are used after specific convolutional layers (e.g., after the third, fourth, and fifth convolutional layers). The last max-pooling layer is followed by a flattening layer (Flatten) (see Figure 10).

As mentioned before, this model has several convolutional layers, allowing it to effectively capture complex hierarchical characteristics in the input images. MaxPooling layers aid in spatial down-sampling, decreasing spatial dimensions and computational effort while preserving crucial features. By applying batch normalization to intermediate feature maps following particular convolutional layers, the training process becomes more stable and converges at a faster rate.

The main objective of this model is to evaluate the effectiveness of training a model that has not been previously trained, using a dataset that comprises thermal images. The training parameters of this model were selected to closely resemble those of previously trained models.

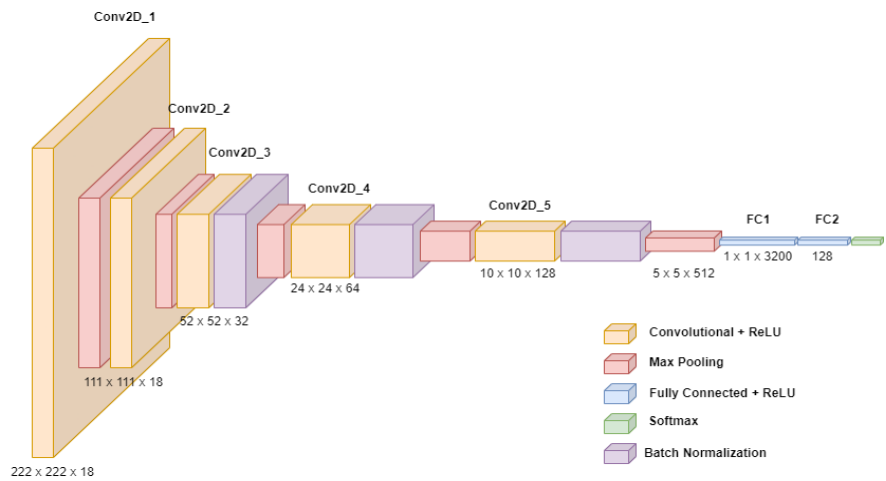

Figure 10. The architecture of the proposed Convolutional model

### 2.5.3. Models Fine-Tuning

The aforementioned models employ varying picture scaling throughout the training phase. The Inception, ResNet, and VGG19 models were trained by downsizing the image dimensions to 224 × 224. However, the Xception model was constructed using images that were upscaled to dimensions of 299 × 299. Most of the training parameters remained unchanged throughout the experiments. Specifically, the learning rate was maintained at a constant value of 0.001, the optimizer used was Adam [21], and the number of training epochs was consistently set to 50 epochs for each architecture that was examined (See Table 4).

Table 5 displays the fine-tuning parameters of the models utilized in this research.

|  | Steps Per Epoch | Learning Rate | Batch Size | Optimizer | Epochs |
|---|---|---|---|---|---|
| **Parameters** | 15 | 0.001 | 32 | Adam | 50 |

# 3. Results and Discussion

The performance results of VGG19, ResNet, InceptionV3, Xception, and the proposed CNN model in detecting honey adulteration are shown in Table 6 and Figures 11,12,13,14,15. The findings demonstrate that our model surpasses the other pretrained models, with Inception V3 appearing as the most exceptional model among these state-of-the-art models. ResNet exhibits the lowest performance among the pretrained models, with minimal levels of accuracy, precision, recall, and substantial loss. The VGG, Inception, and Xception architectures outperform ResNet in capturing distinct patterns or features in thermal images. This superiority is due to the fact that ResNet relies on residual connections to address the issue of vanishing gradients and represent the spatial hierarchy in images. Thermal pictures appear to be more effectively depicted via sequential patterns, such as those employed in VGG, or by factorized convolutions, as used by Inception and Xception. Furthermore, these models exhibit exceptional performance while operating with a restricted amount of data.

On the other hand, the effectiveness of our suggested model can be elucidated by acknowledging that the pretrained models (VGG, Inception, ResNet and Xception) were initially trained on the ImageNet dataset, which consists of a wide variety of images that may substantially differ from thermal images in terms of content, characteristics, and patterns. The intrinsic disparity between the source dataset (ImageNet) and our target dataset of thermal images can result in a misalignment in feature representations. Thermal images frequently display clear thermal patterns, temperature fluctuations, and distinctive visual attributes that may not be accurately represented by models trained on datasets mostly consisting of visible light images.
Our proposed model underwent initial training and fine-tuning using a dataset of thermal images. This enabled the model to use the distinctive features and patterns observed in thermal images and to adapt more effectively to the subtle nuances of thermal information, outperforming models that were pretrained on more general picture datasets like ImageNet.

Moreover, the notable benefit of our model is in its simplicity and lightweight nature in comparison to the other investigated models. This model achieves a more accessible implementation by reducing complexity, resulting in a lower demand for processing resources while maintaining impressive accuracy. This feature has the potential to be very important in real-world situations, as it provides a more effective and economical alternative for implementing on a big scale, especially in the honey industry.

In addition, the distinctive aspect of our work lies in our utilization of honey that has been contaminated with different concentrations (10%, 25%, and 50%). Interestingly, we observe that variations in the amount of honey do not significantly affect the performances of almost all models.
The models' ability to withstand differing amounts of contaminated honey demonstrates their strength in effectively classifying and distinguishing between authentic and adulterated samples. This finding is important because it indicates that the models have the ability to reliably predict outcomes for different honey compositions, showcasing their potential value in real-world scenarios where variations in product quality may occur.

Table 6. displays the performances achieved by different architectures.

| Architecture | Accuracy | Precision | Recall | Loss |
|---|---|---|---|---|
| **InceptionV3** | 0.986 | 0.986 | 0.986 | 0.043 |
| **VGG19** | 0.965 | 0.976 | 0.965 | 0.113 |
| **ResNet50** | 0.881 | 0.855 | 0.965 | 0.274 |
| **Xception** | 0.968 | 0.959 | 0.959 | 0.093 |
| **Our proposed model** | 0.993 | 1.000 | 0.988 | 0.037 |

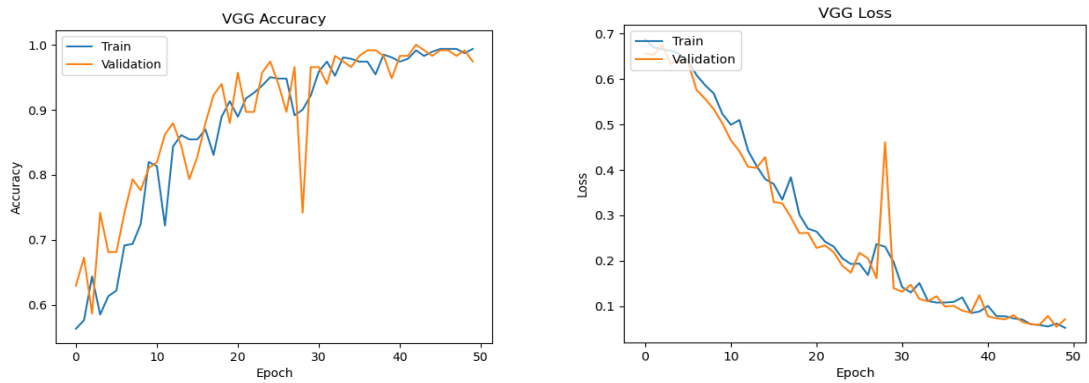

Figure 11. Depicts the accuracy and loss of VGG19 on the training and validation set.

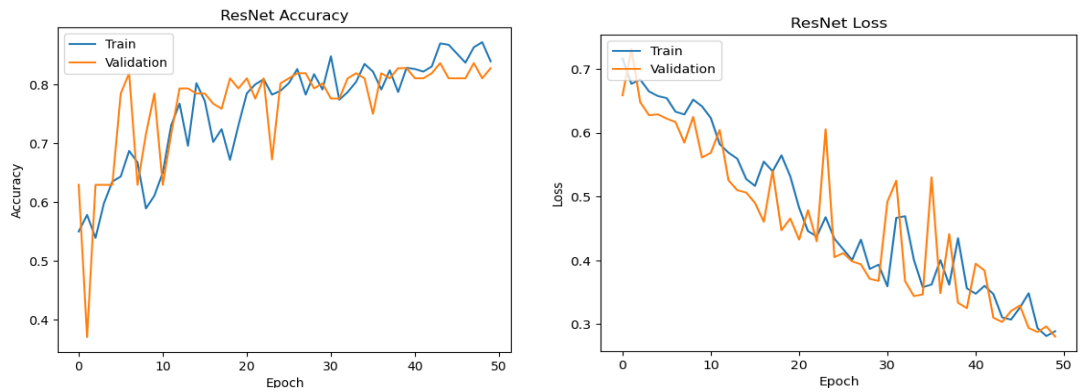

Figure 12. Depicts the accuracy and loss of Resnet on the training and validation set.

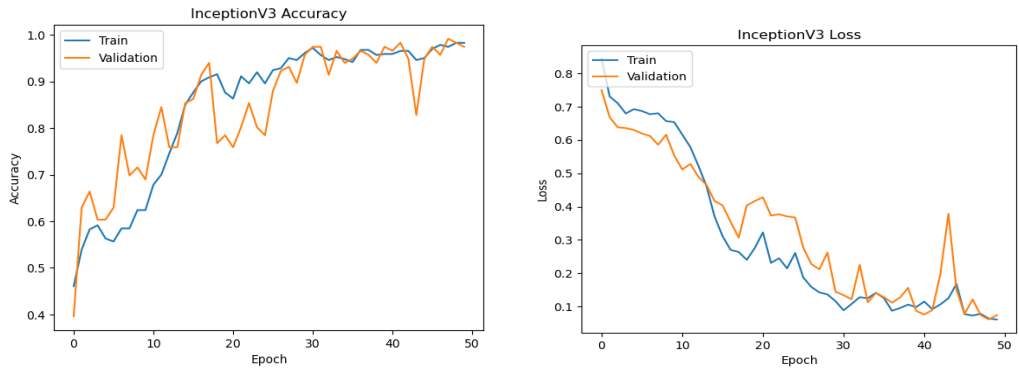

Figure 13. Depicts the accuracy and loss of Resnet on the training and validation set

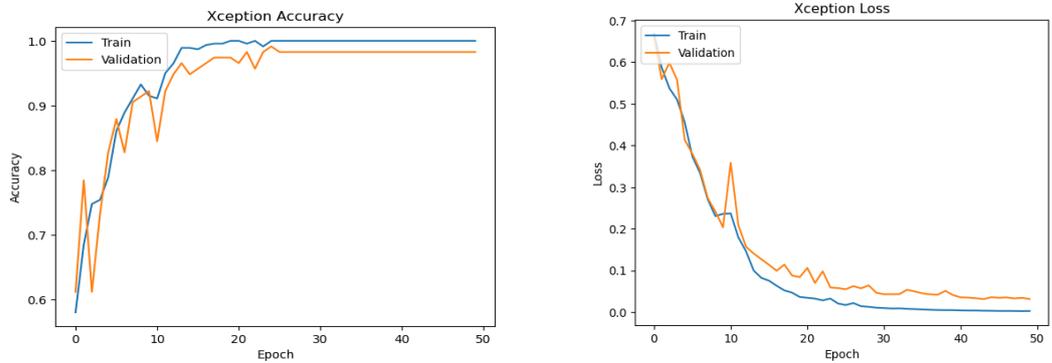

Figure 14. Depicts the accuracy and loss of Xception model on the training and validation set.

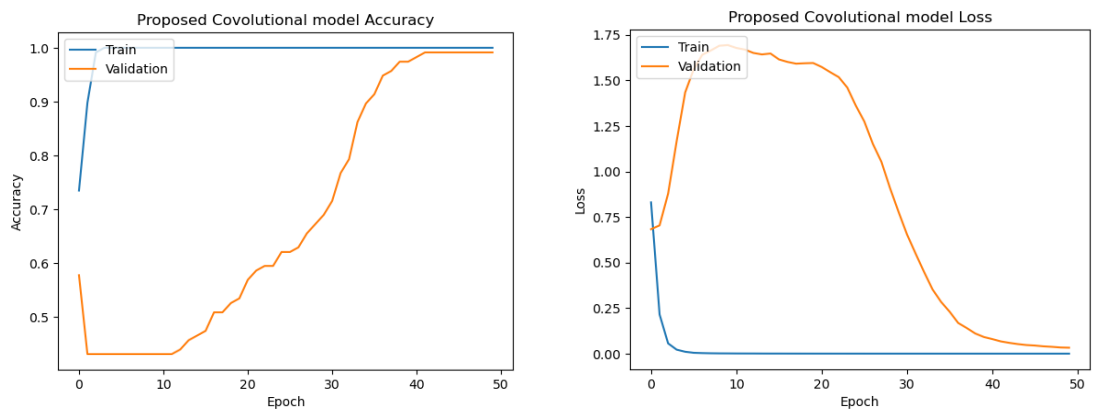

Figure 15. Depicts the accuracy and loss of the proposed model on the training and validation set.

## 4. Conclusion

This study has thoroughly investigated the issue of honey adulteration by employing an innovative approach that utilizes thermal image categorization using several convolutional neural network

(CNN) architectures. The results obtained from well-established models like VGG19, ResNet, InceptionV3, and Xception demonstrated robust performance, confirming their capacity to distinguish between pure and adulterated honey samples with varying levels of contamination. However, the proposed architecture, characterized by its simplicity, outperformed these trained models by achieving an exceptional accuracy of 99%, precision of 100%, recall of 0.988, and loss of 0.037.

The proposal's practical benefit arises from its direct execution and superior operational effectiveness, making it a viable option for extensive quality monitoring in the honey industry. The simplification of the methodology has not compromised its ability to detect spoilage, offering a cost-effective and efficient approach to verifying the authenticity of bee products.

Possible expansions of this study involve investigating novel convolutional neural network (CNN) structures tailored to specific circumstances in the honey industry, as well as other varieties of honeys. Additionally, incorporating alternative imaging modalities could lead to a more comprehensive analysis.

Potential future endeavors may include expanding this approach to additional food items, verifying its effectiveness using larger and more varied datasets, and investigating the incorporation of artificial intelligence into other components of the food distribution network. Overall, this work provides a noteworthy contribution by integrating technology and food quality, thereby facilitating promising advancements in food safety and safeguarding consumer interests.

- **Conflicts of Interest**: The authors declare no conflict of interest.
- **Funding**: The authors are grateful to the funding of the CNRST and Mohammed VI Polytechnic University.

## Authors

**Boulbarj Ilias**. is now pursuing a Ph.D. at the laboratory "Image and Recognition of Forms – Intelligent and Communicating Systems" at the Faculty of Sciences in Agadir, Morocco. He obtained a master's degree in Applied Numerical and Statistical Method (MNSA) from the faculty of Sciences Aïn-Chock at Hassan II University. Mr. Boulbarj's research focuses on Machine Learning, specifically in the field of Computer Vision.

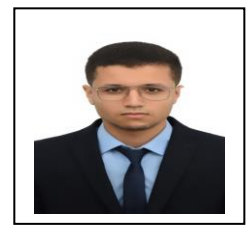

**Abdelaziz Bouklouze** is a research professor in the Faculty of Medicine and Pharmacy, University Mohammed V, Rabat, Morocco and Operations Research and the head of Biopharmaceutical and Toxicological Analysis Research Team, Research Chair in analytical techniques (targeted and untargeted) in tandem with chemometrics and machine learning tools for quality control in Pharmaceutical, Environmental, Food and Clinical fields.

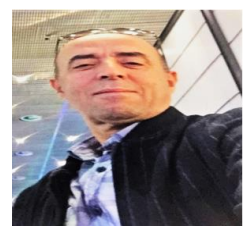

**Yousra Alami** is a doctoral candidate in her fourth year of study at the Research Team for Biopharmaceutical and Toxicological Analyses laboratory, located at the Faculty of Medicine and Pharmacy in Rabat, Morocco. Proficient in the field of analytical chemistry, with expertise in conducting physicochemical analysis of health goods and ensuring the quality control of pharmaceuticals.

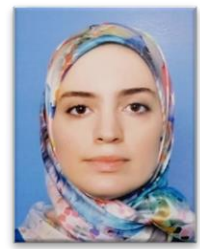

**Samira Douzi** holds the position of associate professor in the Faculty of Medicine and Pharmacy, University Mohammed V, located in Rabat, Morocco. She obtained her PhD in Deep Learning systems for cyber security from the University of Mohammed V Rabat. She authored numerous research publications mostly focused on computer vision applications for illness detection, particularly in the field of cancer.

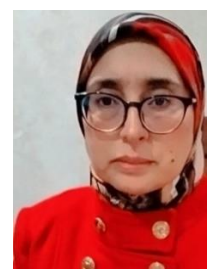

**Hassan Douzi** is a research professor in mathematics at Ibn Zohr University, Agadir, Morocco. He received his PhD in "Mathematics for decision" from the University of Paris IX Dauphine 1992. He published several research papers, essentially in Applied mathematics related to image processing.

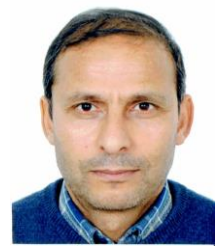